\documentclass[nohyperref]{article}
\pdfoutput=1
\usepackage{microtype}
\usepackage{graphicx}
\usepackage{subfigure}
\usepackage{booktabs} 

\usepackage[hyphens]{url}
\usepackage{hyperref}
\usepackage[hyphenbreaks]{breakurl}

\usepackage[accepted, nohyperref]{icml2023}

\usepackage{amsmath}
\usepackage{amssymb}
\usepackage{mathtools}
\usepackage{amsthm}

\usepackage{amsmath,amsfonts,bm}

\def\eqref#1{equation~\ref{#1}}

\def\1{\bm{1}}

\def\vy{{\bm{y}}}

\def\mC{{\bm{C}}}

\def\mH{{\bm{H}}}

\def\mL{{\bm{L}}}
\def\mM{{\bm{M}}}

\def\mR{{\bm{R}}}
\def\mS{{\bm{S}}}

\def\mX{{\bm{X}}}
\def\mY{{\bm{Y}}}

\DeclareMathAlphabet{\mathsfit}{\encodingdefault}{\sfdefault}{m}{sl}
\SetMathAlphabet{\mathsfit}{bold}{\encodingdefault}{\sfdefault}{bx}{n}

\newcommand{\R}{\mathbb{R}}

\usepackage[capitalize,noabbrev]{cleveref}

\theoremstyle{plain}

\theoremstyle{definition}

\theoremstyle{remark}

\usepackage[textsize=tiny]{todonotes}

\icmltitlerunning{Shiftable Context: Addressing Training-Inference Context Mismatch in Simultaneous Speech Translation}

\begin{document}

\twocolumn[
\icmltitle{Shiftable Context: Addressing Training-Inference Context Mismatch in Simultaneous Speech Translation}



\icmlsetsymbol{equal}{*}

\begin{icmlauthorlist}
\icmlauthor{Matthew Raffel}{OSU}
\icmlauthor{Drew Penney}{OSU}
\icmlauthor{Lizhong Chen}{OSU}
\end{icmlauthorlist}

\icmlaffiliation{OSU}{School of Electrical Engineering and Computer Science, Oregon State University, Corvallis, OR, United States}

\icmlcorrespondingauthor{Matthew Raffel}{raffelm@oregonstate.edu}
\icmlcorrespondingauthor{Drew Penney}{penneyd@oregonstate.edu}
\icmlcorrespondingauthor{Lizhong Chen}{chenliz@oregonstate.edu}

\icmlkeywords{Machine Learning, ICML, Simultaneous Translation, SimulST, Transformer, Speech}

\vskip 0.3in
]



\printAffiliationsAndNotice{}  
\begin{abstract}
Transformer models using segment-based processing have been an effective architecture for simultaneous speech translation.  However, such models create a context mismatch between training and inference environments, hindering potential translation accuracy.  We solve this issue by proposing \textit{Shiftable Context}, a simple yet effective scheme to ensure that consistent segment and context sizes are maintained throughout training and inference, even with the presence of partially filled segments due to the streaming nature of simultaneous translation.  Shiftable Context is also broadly applicable to segment-based transformers for streaming tasks.  Our experiments on the English-German, English-French, and English-Spanish language pairs from the MUST-C dataset demonstrate that when applied to the Augmented Memory Transformer, a state-of-the-art model for simultaneous speech translation, the proposed scheme achieves an average increase of 2.09, 1.83, and 1.95 BLEU scores across each wait-$k$ value for the three language pairs, respectively, with a minimal impact on computation-aware Average Lagging.

\end{abstract}
\section{Introduction}
Simultaneous speech-to-text translation (SimulST) aims to produce an output text translation concurrently with an oncoming speech input.  Performing accurate simultaneous translation over long periods for humans is extremely difficult due to the immense strain placed on the brain.  Machine learning is promising to help fill the role but also faces significant challenges, primarily because translation must be generated with partial input sentences, and computation must be done in a real-time fashion during inference.  Given the broad potential applications of automated simultaneous translation, there is a pressing need to develop highly accurate and highly efficient SimulST models.

While earlier transformer-based SimulST models take the received speech directly as the input, recent models based on segments have demonstrated better performance with less computation \cite{dong2019self, dai2019transformer,ma2021streaming,wu2020streaming,shi2021emformer}.  These segment-based transformers break an input sequence into segments and sequentially process each segment individually before concatenating them together in the encoder.  
To retain some of the information in prior segments, additional left context or summarization tokens can be provided to the current segment that is being processed.  As self-attention is calculated only within the segment (including the additional context), complexity is greatly reduced.  The latest work along this line is the Augmented Memory Transformer (AMT) \cite{wu2020streaming, ma2021streaming}, where each segment overlaps with previous and subsequent segments with a left and a right context, respectively, which also alleviates the issue of word boundaries.

Even though Augmented Memory Transformer has state-of-the-art performance for SimulST, we found that it suffers from a context mismatch issue between its training and inference environments.  Specifically, during inference, each of the left, center, and right contexts of a segment can be partially filled under different scenarios, which deviates from the fixed segment size assumed in training.
This mismatch issue occurs frequently and exists in other segment-based transformers (such as Transformer-XL \cite{dai2019transformer}, Emformer \cite{shi2021emformer}, and Implicit Memory Transformer \cite{raffel-etal-2023-implicit}) for streaming tasks in general.

To address this issue, we propose \textit{Shiftable Context}, which includes multiple techniques to produce consistent segment sizes for better alignment of training and inference in streaming tasks for segment-based transformers.  This is achieved by proposing a shiftable left, center, and right context.  In the case of the shiftable center and right context, when a segment does not reach the assumed size of tokens, they are instead mapped to the left context, increasing its size.  Similarly, for the shiftable left context, when a segment does not have prior information from the input sequence, the tokens normally associated with the left context are remapped to create a larger right context.  

We trained and evaluated our models on the English-German, English-French, and English-Spanish language pairs from the MUST-C data set \cite{CATTONI2021101155}.  The evaluation metrics utilized to determine the efficacy of any changes made were computation-aware Average Lagging, a latency metric \cite{ma2020simulmt}, and BLEU score, a translation accuracy metric \cite{Papineni2002BleuAM}, using SimulEval \cite{ma-etal-2020-simuleval}. Our evaluation results on English-German translation demonstrated that, when applied individually to the tst-COMMON test set, the proposed technique of shiftable left, center, and right context improved the BLEU score of Augmented Memory Transformer by 0.42, 0.66, and 1.10 on average. When combined together, the proposed shiftable context approach achieved an average of 2.09  BLEU increase across all wait-$k$ values with a minor increase in computation-aware Average Lagging. Equally impressive results were found in English-French and English-Spanish translations yielding average BLEU score increases of 1.83 and 1.96 BLEU, respectively. 

The main contributions of this paper are:
\vspace{-0.2cm}
\begin{enumerate}
    \item Identified a mismatch between training and inference for segment-based transformers for SimulST. 
    \vspace{-0.2cm}
    \item Proposed a novel and broadly applicable approach to any segment-based transformer, which eliminates the training-inference context mismatch for streaming tasks like SimulST by employing a shiftable left, center, and right context.
    \vspace{-0.2cm}
    \item Demonstrated the efficacy of the proposed techniques by conducting extensive experiments that evaluated techniques individually and collectively with multiple language pairs and many wait-$k$ values.
\end{enumerate}

\section{Background and Related Work}
\label{Sec:Background}
\subsection{Simultaneous Translation}
\textbf{Wait-$k$ Policy:}
The wait-$k$ policy was first introduced for simultaneous text translation (SimulMT).  The policy is simple yet effective where the decoder will wait for $k$ encoder output tokens associated with the first $k$ words in a sentence before beginning the translation \cite{ma2018stacl}.  The model performing the translation then alternates between producing a target word and reading a new source word.

\textbf{Fixed Pre-decision Module:} 
The fixed pre-decision module bridges the gap between SimulMT and SimulST.  Prior to its inception, simultaneous policies such as the wait-$k$ policy were inapplicable to SimulST, as the granularity of input tokens for speech was too fine-grained \cite{ma2020simulmt}.  The fixed pre-decision module solves this issue by grouping encoder states into chunks before providing them to the decoder.  The size of these chunks is determined by a hyperparameter called the pre-decision ratio, which specifies the exact number of encoder output tokens to provide to the decoder at each time step.

\textbf{Simultaneous Decoder:}
A simultaneous decoder is a slight modification to the decoder introduced in \cite{vaswani2017attention} that allows it to perform SimulST by making use of the wait-$k$ policy and a fixed pre-decision module \citep{ma2021streaming}.  In the simultaneous decoder, the pre-decision module separates the incremental speech tokens into chunks of a fixed size.  Each fixed-sized chunk is treated akin to a word in SimulMT.  Therefore, by applying a pre-decision module to the decoder, the wait-$k$ policy is applicable for SimulST tasks.  As such, the simultaneous decoder will wait for $k$ fixed-size chunks before beginning translation.  Once translation begins, the simultaneous decoder will alternate between reading and writing new inputs and outputs at the per-chunk granularity.

\subsection{Augmented Memory Transformer}
\begin{figure}[ht]
\vskip 0.1in
\begin{center}
\centerline{\includegraphics[width=\columnwidth]{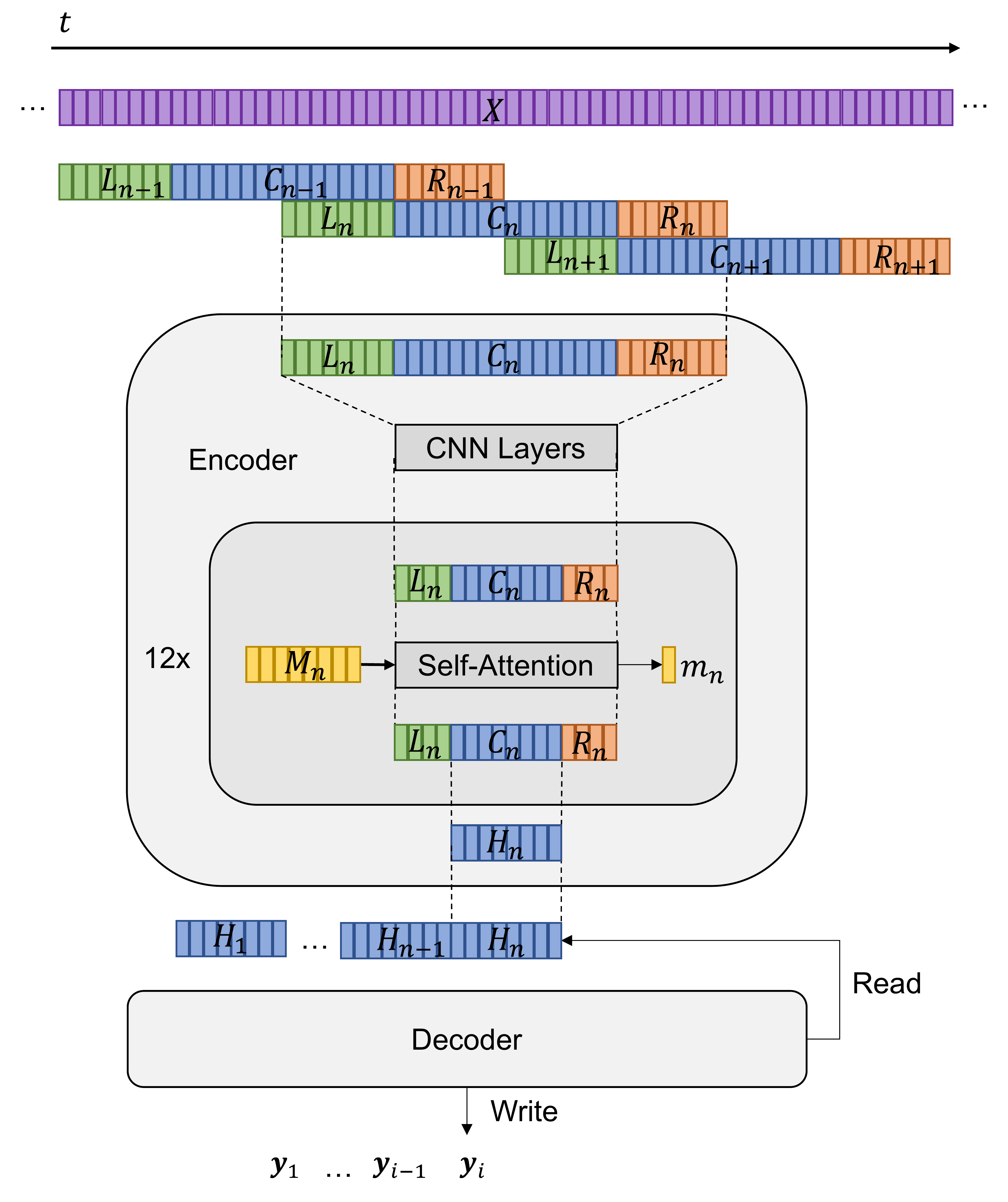}}
\caption{An example of the Augmented Memory Transformer encoder breaking an input sequence, $\mX$, into segments, $\mS_n$, individually processing them, concatenating the outputs, and passing the concatenated output to the simultaneous decoder.}
\label{fig:augmented}
\end{center}
\vskip -0.2in
\end{figure}

The encoder in the Augmented Memory Transformer breaks a speech input sequence $\mX = [\mS_1, \mS_2,...\mS_n,...]$ into segments $\mS_n \in \R^{s\times d}$, where $s$ is the size of the segment, and passes the results sequentially to subsampling convolution layers.  Each segment consists of a concatenation of a left context $\mL_n \in \R^{l\times d}$, a center context $\mC_n \in \R^{c\times d}$, and a right context $\mR_n\in \R^{r\times d}$ of sizes $l$, $c$, and $r$, respectively.  Each segment can be represented by the following equation:
\begin{equation}
    \mS_n = [\mL_n,\mC_n, \mR_n]
    \label{eq:segment}
\end{equation}
Equation \ref{eq:segment} uses [.] as a concatenation notation to concatenate the left, center, and right context together.  As such, $s=l+c+r$.  The $n$ subscript denotes the position of the segment in the entire sequence.  Figure \ref{fig:augmented} represents the architecture of the Augmented Memory Transformer, which processes each segment sequentially.  A segment overlaps with the previous and subsequent segments using the left and right context.  As a result, the model is able to reduce disrupting word boundaries.  Since each segment is passed through the encoder sequentially, the self-attention calculation is at the segment level.  For self-attention, the queries are created from Equation \ref{eq:segment} (after subsampling) concatenated with a summarization query, $\sigma_n$.  The attention output associated with this summarization query creates a single memory bank, $m_n$, that summarizes the current segment.  The keys and the values are augmented with $N$ additional memory banks, $\mM_n \in \R^{N\times d}$, that summarize the $N$ previous segments.
After each segment passes through the encoder, the auxiliary left and right contexts are stripped off, and only the center contexts are concatenated and passed to the simultaneous decoder.  During simultaneous translation, the simultaneous decoder alternates between reading encoder states and writing the translation after adhering to the wait-$k$ policy.

\section{Training-Inference Context Mismatch}

In this section, we describe the context mismatch issue that we have identified for segment-based transformers when performing simultaneous translation tasks. 

Based on the previous section, except for starting and terminating segments, each segment contains a fixed number of chunks.  For example, in Figure \ref{fig:augmented}, if the left, center, and right context of a segment has 8, 16, and 8 tokens, respectively, and a chunk has 4 tokens, then a segment $\mS_n$ consists of 32 tokens or 8 chunks.  As simultaneous translation is real-time, the encoder self-attention is triggered after each new chunk (e.g., a word, conceptually) is received, so a new encoder hidden state can be provided to the simultaneous decoder to generate an output word.  Therefore, each new chunk arrival triggers a self-attention calculation among all the chunks that have been received in the current center context, along with the left context (and right context if applicable).  For instance, the arrival of the first chunk of a segment $\mS_n$ triggers the attention calculation among 12 tokens (8 tokens from the left context + 4 tokens from the chunk) to generate an intermediate hidden state $\mH_n$ of the segment for the decoder, the arrival of the second chunk of the segment triggers the attention calculation among 16 tokens (8 tokens from the left context + 8 tokens from first two chunks), and so on.  The hidden state $\mH_n$ of the segment is continuously recomputed and provided to the decoder until it has processed a complete 32 tokens.

In practice, recomputing the same segment after each new chunk arrival is very costly.  Since the entire input sequence is available during training, a more practical and efficient implementation such as the one used by Augmented Memory Transformer is to simply break the entire input sequence into complete segments and calculate the self-attention of each complete segment once, as if all the chunks of a segment arrive at the same time.  This removes the need to reprocess the same segment multiple times, leading to a substantial saving in training time.  Furthermore, such an implementation is more parallelizable as there are fewer sequential computations.  As a result of the parallelized training, the encoder provides the decoder with a single collection of hidden states created to represent the entire sequence.  To still correctly model simultaneous translation, the encoder-decoder attention calculation only attends to a portion of the encoder's hidden states for each prediction.

While such a training implementation is faster and less computationally expensive, the model is trained to perform best when generating hidden states under a fixed segment size (e.g., the complete 32 tokens) and a fixed center context size (e.g., 16 tokens). However, during inference for simultaneous translation, the number of available tokens used to generate hidden states varies depending on the actual number of chunks that have been received.  This creates a training-inference context mismatch which can be problematic as the decoder is not trained to handle intermediate hidden states well. It is worth noting that this mismatch occurs very frequently during inference, except for those chunks that happen to arrive at segment boundaries. To address this important issue, we need a new way to align the context used for hidden state generation without influencing the already efficient training process. 

\section{Methods: Shiftable Context}
In this work, we propose \textit{Shiftable Context}, a simple yet effective scheme to remove the training-inference context mismatch for segment-based transformers. As a context mismatch may occur in the left, center, or right context of a segment, the proposed scheme consists of three techniques, namely shiftable center, right, and left context, to ensure that the inference of simultaneous translation retains consistently sized segments.  In what follows, we first present each of these techniques individually before finally combining them into the complete scheme. 

\subsection{Shiftable Center Context}
\begin{figure}[ht]
\vskip 0.1in
\begin{center}
\centerline{\includegraphics[width=0.85\columnwidth]{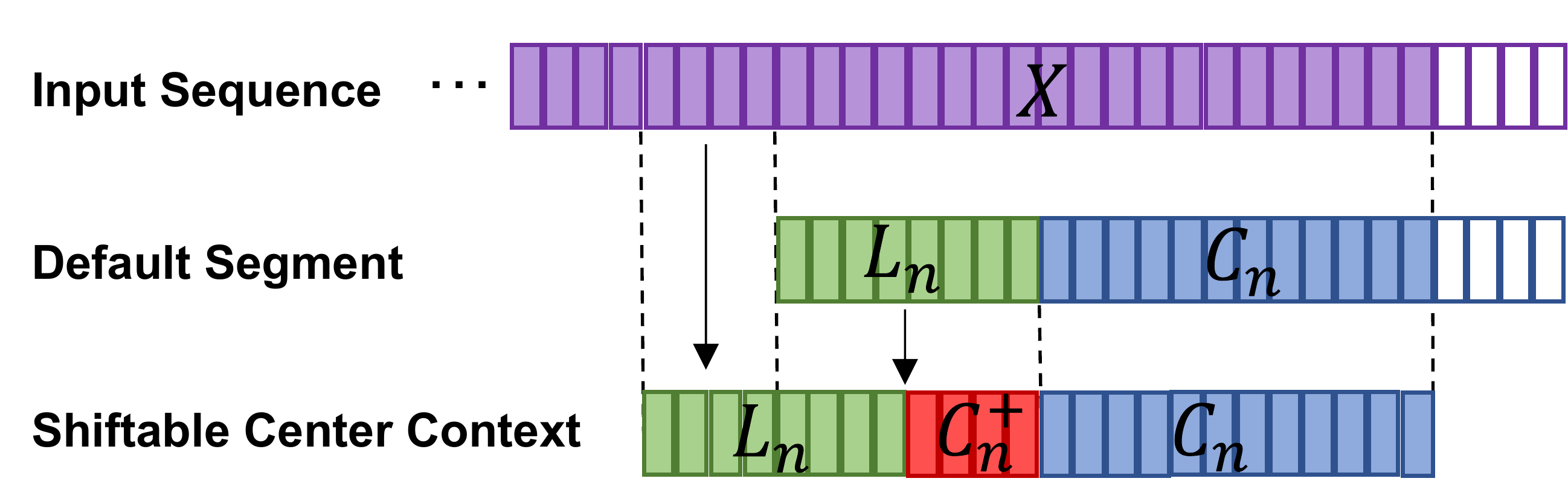}}
\caption{An example demonstrating the default method of creating a segment from the input sequence, $\mX$, where there is an insufficient number of tokens to fill the center context, $\mC_n$, alongside the proposed segment formation employing a shiftable center context, $\mC_n^+$, to maintain consistently sized center context.}
\label{fig:center}
\end{center}
\vskip -0.2in
\end{figure}

Our first technique, shiftable center context, targets the mismatch that occurs in the center context of a segment. During simultaneous translation inference, the transformer encoder continuously receives partial inputs.  If the partial input sequence is not divisible by the size of the center context, then the center context is only partially filled.  Figure \ref{fig:center} depicts such an example where $\mX$ denotes the partial input sequence. The complete center context would normally have 16 tokens, but in our example, it is currently filled with 12 tokens (equivalently 3 chunks).  Since the third chunk has recently arrived, the self-attention needs to be calculated for it using the partial center context (to generate an intermediate hidden state as discussed in Section \ref{Sec:Background}). This creates a mismatch as training assumed a fixed center context contradicting the partial center context seen in inference.

To address this mismatch, the proposed shiftable center context utilizes additional tokens from previous segments to ensure that the size of the center context remains constant.  This is demonstrated in the bottom segment of Figure \ref{fig:center}.  In the example, the shiftable center context, $\mC_n^+$, is composed of the last 4 tokens of the left context in the default segment.  As a result, the new left context is instead created from the first 4 tokens of the default segment's left context plus 4 additional tokens from the input sequence.  Essentially, the left context is shifted left to make space for a larger center context.  If generalized, the partially filled segment can be represented by Equation \ref{eq:center}:    
\begin{equation}
    \mS_n = [\mL_n, \mC_n^+, \mC_n]
    \label{eq:center}
\end{equation}

In Equation \ref{eq:center}, $\mC_n^+$ refers to the shiftable center context.  The subscript $n$ indicates the segment position in the entire sequence.  Each instance of $\mC_n^+$ can vary in size between 0 and $c-1$ to ensure that the combined new center context reaches the fixed center context size each time a new chunk is received.  
Note that no right context is included in the equation.  This is because if there are not sufficient tokens to completely fill the center context, the right context must be empty.  The next subsection discusses how mismatches are handled if the center context is completely filled, but the right context is not.

\subsection{Shiftable Right Context}
\begin{figure}[ht]
\vskip 0.1in
\begin{center}
\centerline{\includegraphics[width=0.95\columnwidth]{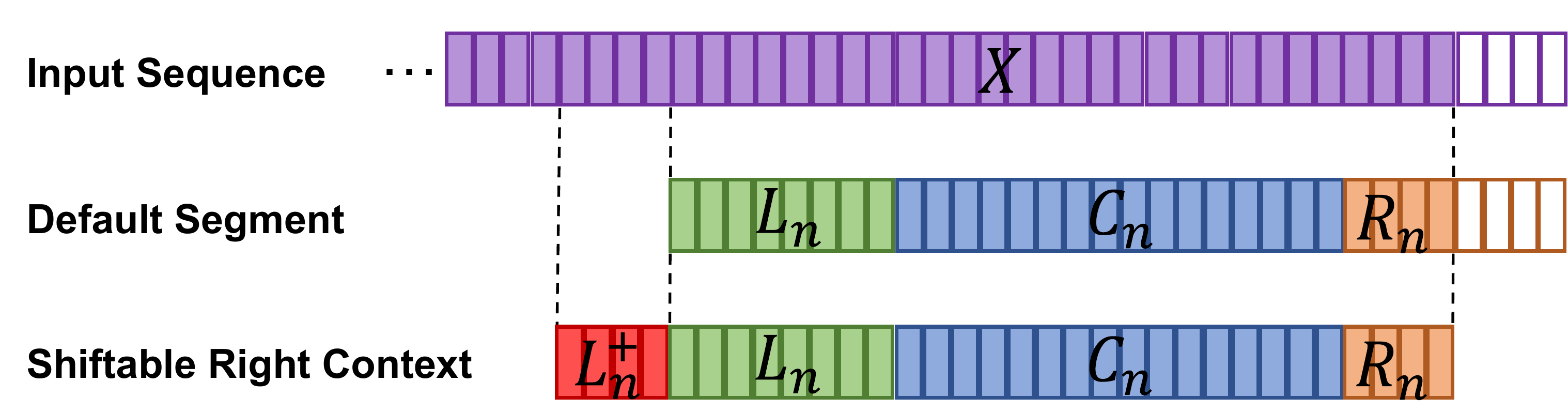}}
\caption{An example demonstrating the default method of creating a segment from the input sequence, $\mX$, where there is an insufficient number of tokens to fill the right context, $\mR_n$, alongside the proposed shiftable right context to remap tokens from the right context to the left context to keep consistently sized segments.}
\label{fig:right}
\end{center}
\vskip -0.2in
\end{figure}

Our second technique, shiftable right context, targets the mismatch that is caused by a partially filled right context during inference. Figure \ref{fig:right} illustrates an example where the partial input sequence $\mX$ is 4 tokens short of filling up the right context.  Consequently, the entire segment is only partially filled, which does not match the fixed segment size the model was trained to use to generate encoder hidden states accurately.  
 
This is addressed by the proposed shiftable right context, which remaps the space of unfilled tokens in the right context to the left context. We demonstrate this solution in the bottom segment of Figure \ref{fig:right}.  In the example, the 4 unfilled token spaces in the right context are repurposed as an additional left context $\mL_n^+$, using tokens in $\mX$ that are prior to the left context in the default segment.  As such, the size of the entire segment maintains at 32 tokens rather than 28 tokens.

In general, we can represent each partially filled segment with the inclusion of a shiftable right context 
with Equation \ref{eq:right}:
\begin{equation}
    \mS_n = [\mL_n^+,\mL_n, \mC_n]
    \label{eq:right}
\end{equation}
In Equation \ref{eq:right}, $\mL_n^+$ represents the additional left context enabled by shifting the position of the right context, ultimately including more information from the previous segment.  The size of $\mL_n^+$ is at most $r$.  By doing so, the number of tokens in the entire segment reaches its fixed length of $l+c+r$, as long as additional tokens are available from the partial input sequence and the center context is complete. 

\subsection{Shiftable Left Context}
\label{Sec:ShiftableLeftContext}
\begin{figure}[ht]
\vskip 0.1in
\begin{center}
\centerline{\includegraphics[width=\columnwidth]{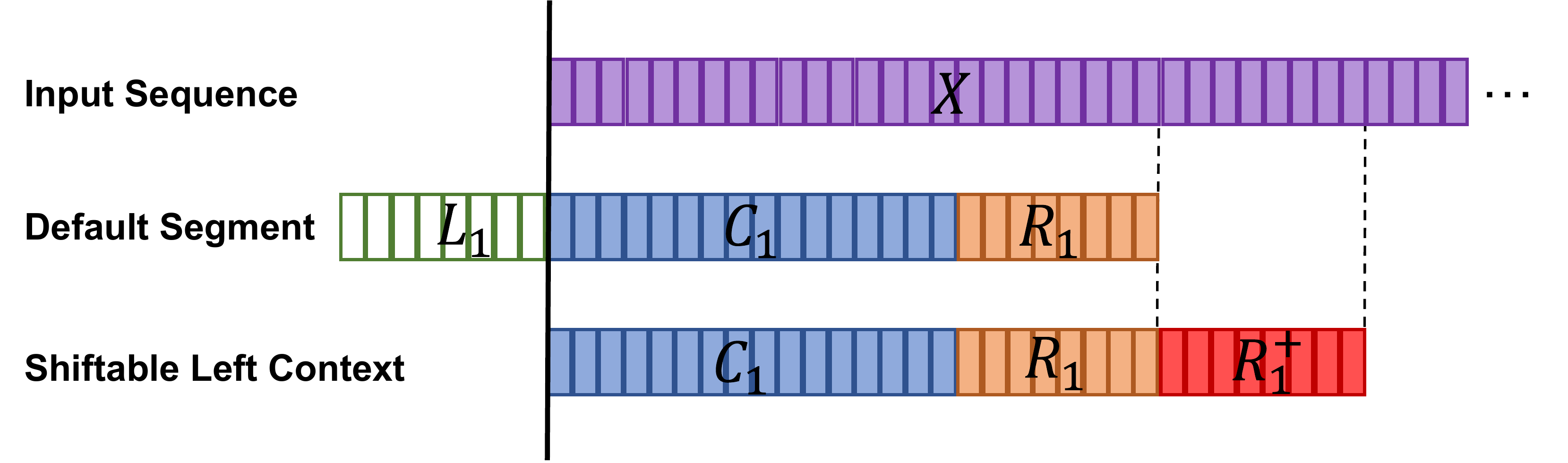}}
\caption{An example demonstrating the default method of creating a segment from the input sequence, $\mX$, where there are no tokens to fill the left context, $\mL_n$, alongside the proposed shiftable left context to remap tokens from the left context to the right context to keep consistently sized segments.}
\label{fig:left}
\end{center}
\vskip -0.2in
\end{figure}

Whenever the transformer begins simultaneous translation, there is no prior contextual information.  Therefore, the first segment will never have a left context.  Since an overwhelming majority of the segments that the Augmented Memory Transformer is trained on are composed of a completed segment of size $l+c+r$, it is not suited to interpret segments of size $c+r$.  This issue is shown in Figure \ref{fig:left}, whereby the first tokens of input $\mX$ are directly used as the center context since there are no prior tokens to use as the left context.  As a result, there are 8 left context tokens missing that are normally available.  Even though such an issue occurs once per sentence, the representation of the first segment is critical to the translation since the first couple of predicted outputs for the decoder are heavily influenced by its hidden state.  This is due to the lack of hidden states generated at the beginning of sentence translation and the decoder reusing its predicted output for its own self-attention calculation.  As such, if the first prediction for the decoder is incorrect/inaccurate, it will steer the decoder to create wrong predictions during translation continuously (our ablation study in Section \ref{Sec:Ablation} confirms that improving the translation of the first segment results in a sizable impact on the overall BLEU score).   

Our solution is to adopt a shiftable left context which remaps the position of the unfilled left context tokens to create a greater right context.  As a result, the right context has a size of $l+r$, and the entire segment reaches its fixed size of $l+c+r$.  This technique is illustrated in Figure \ref{fig:left}, where the 8 tokens that would normally be used as left context are instead used to create an extended right context of 16 tokens.  The equation representation of a segment using a shiftable left context is as follows:
\begin{equation}
    \mS_n = [\mC_n, \mR_n, \mR_n^+]
    \label{eq:left}
\end{equation}

In Equation \ref{eq:left}, $ \mR_n^+$ represents the additional right context enabled by shifting the position of the left context.  The maximum size of the shiftable left context is $l$, and it is only applicable if there is an available right context since, without additional right context from the input sequence, there are no tokens to remap the left context.  Such an issue is partially solved by the wait-$k$ policy, which naturally includes a brief waiting period, allowing a greater number of input tokens to arrive and be utilized as the extended right context.  Even without a sufficiently large waiting period, the shiftable left context still becomes useable early in the translation period.  Given the importance of the first segment to SimulST, the shiftable left context provides the opportunity to improve low-latency scenarios by providing a more accurate beginning to a translated sentence.  

\subsection{Putting It Together: Shiftable Context}


The proposed shiftable center, right, and left context techniques complement each other.  By combining the techniques, we are able to ensure that a fixed segment size and a fixed center context size are used when creating the hidden states for the simultaneous decoder, even when partial chunks are received at each time step. 

If we were to include the shiftable center, right and left contexts in the representation of a segment, it could be represented with the following equation: 
\begin{equation}
    \mS_n = [\mL_n^+,\mL_n,\mC_n^+, \mC_n, \mR_n, \mR_n^+]
    \label{eq:complete}
\end{equation}

None of the terms in Equation \ref{eq:complete} share tokens, and the total size of such a segment is $l+c+r$ whenever possible, which is the normal size of each segment during training.  Note that, not all techniques are activated concurrently.  For instance, since the shiftable center and right context require tokens from previous segments when activated, it is not activated if the shiftable left context is in use (implying no left context available).  
In general, (1) the shiftable left context is activated when processing the first segment; (2) the shiftable center context is activated when the center context is partially filled; and (3) the shiftable right context is activated when the center or right context is partially filled, as an incomplete center context automatically implies an empty right context which can be remapped as the additional left context.


When all these techniques are applied, the computational cost of training the Augmented Memory Transformer remains constant.  During inference, the computational cost of shiftable context may increase slightly, as additional tokens may be added to form complete segments.  However, such an impact is minimal, as presented in the evaluation in Section \ref{Sec:Results}, along with many other results.  We provide token-level simultaneous translation examples comparing an Augmented Memory Transformer with and without our shiftable context technique in Appendix \ref{Appendix:Examples}.
  
\section{Experimental Setup}
\subsection{Data Set}
We conducted experiments on the English-German (en-de), English-French (en-fr), and English-Spanish (en-es) language pairs from the MUST-C dataset \citep{CATTONI2021101155}. The data preparation scripts for the MUST-C dataset are provided in Fairseq\footnote{\url{https://github.com/facebookresearch/fairseq}} \cite{ott2019fairseq, wang2020fairseqs2t}, whereby Kaldi is used for 80-dimensional log-mel filter bank features, and text is tokenized with a SentencePiece 10k unigram vocabulary.  The training was conducted on the train set.  After each epoch, each model was validated against the dev set.  
 
\subsection{Model Hyperparameters}
 Our Augmented Memory Transformer has 33.1 M parameters.  Its encoder begins with 2 convolution layers with a combined subsampling factor of 4, followed by a feed-forward neural network.  Similarly, the encoders of each consisted of 12 layers, and their decoders consisted of 6 layers.  Each of these layers had a hidden size of 256 with 4 attention heads. Layer normalization was performed prior to each layer.  Additionally, we trained each Augmented Memory Transformer with a wait-1, wait-3, wait-5, and wait-7 policy using a pre-decision ratio of 8 \cite{ma2018stacl, ma2020simulmt}.  Such an approach allowed us to analyze how each of our proposed schemes scaled with latency while also providing more certainty about our results.  The segment of each Augmented Memory Transformer was composed of a left context of 32 tokens, a center context of 64 tokens, and a right context of 32 tokens.  The encoder self-attention calculation used 3 memory banks.  The clipping distance of the relative positional encodings was 16 tokens \cite{shaw2018self}.

\subsection{Training Hyperparameters}
All training was performed on a single V100-32GB GPU.  The training process consisted of ASR pretraining followed by SimulST training.  For the ASR pretraining, the model was trained with label-smoothed cross-entropy loss, the Adam optimizer \cite{kingma2014adam}, and an inverse square root scheduler. Each model was trained with a warmup period of 4000 updates, where the learning rate was 0.0001, followed by a learning rate of 0.0007.  The only regularization for the ASR pretraining was a dropout of 0.1.  Each ASR pretrained model used early stopping with a patience of 5.  As a result, pre-training would stop, and SimulST training would begin if the model did not improve against the validation dev set after 5 epochs.  

For the SimulST training, the model was also trained with label-smoothed cross-entropy loss, the Adam optimizer, and an inverse square root scheduler.  There was a warmup period of 7500 updates where the learning rate of 0.0001, followed by a learning rate of 0.00035.  To regularize the model weights, we used a weight decay value of 0.0001, a dropout of 0.1, an activation dropout of 0.2, and an attention dropout of 0.2.  All models were trained with early stopping using a patience of 10.  After the training was complete, the final 10 checkpoints were averaged. Doing so acted as a regularizer and reduced the odds of randomness interfering with the evaluation results.

\subsection{Evaluation Method}
The translation quality and latency were determined by detokenized BLEU with SacreBLEU \cite{post2018call}, and computation-aware Average Lagging \cite{ma2018stacl, ma2020simulmt}, respectively.  Computation-aware Average Lagging measures how far the target translation is lagging behind the source inputs in milliseconds and takes into account computational time. 
 Equation \ref{eq:AL} provides a mathematical representation of computation-aware Average Lagging.  
\begin{equation}
    \text{AL}=\dfrac{1}{\tau(||\mX||)}\sum_{i=1}^{\tau(||\mX||)}d(y_i)-\dfrac{||\mX||}{||\mY^*||}\cdot T \cdot (i-1)
    \label{eq:AL}
\end{equation}
In Equation \ref{eq:AL}, the input sequence is denoted as $\mX$, where each token represents $T$ milliseconds of information, and the reference translation is denoted as $\mY^*$.  Finally, $\tau(||\mX||)$ represents the index of the first target token generated after the entire source input is read, and $d(\vy_i)$ represents the computational time needed to generate token $\vy_i$. The evaluations were all performed on a single V100-32GB GPU.  The two evaluation sets used to determine the performance of the model were tst-COMMON and tst-HE.  We opted to evaluate using different test sets to provide a more robust confirmation of the efficacy of our schemes.  Both the BLEU score and computation-aware Average Lagging were obtained using the SimulEval toolkit\footnote{\url{https://github.com/facebookresearch/SimulEval}}, which simulates simultaneous speech translation \citep{ma-etal-2020-simuleval}.
\section{Results}
\label{Sec:Results}
\subsection{Main Results}
\begin{figure}[ht]
\vskip 0.1in
\begin{center}
\centerline{\includegraphics[width=\columnwidth]{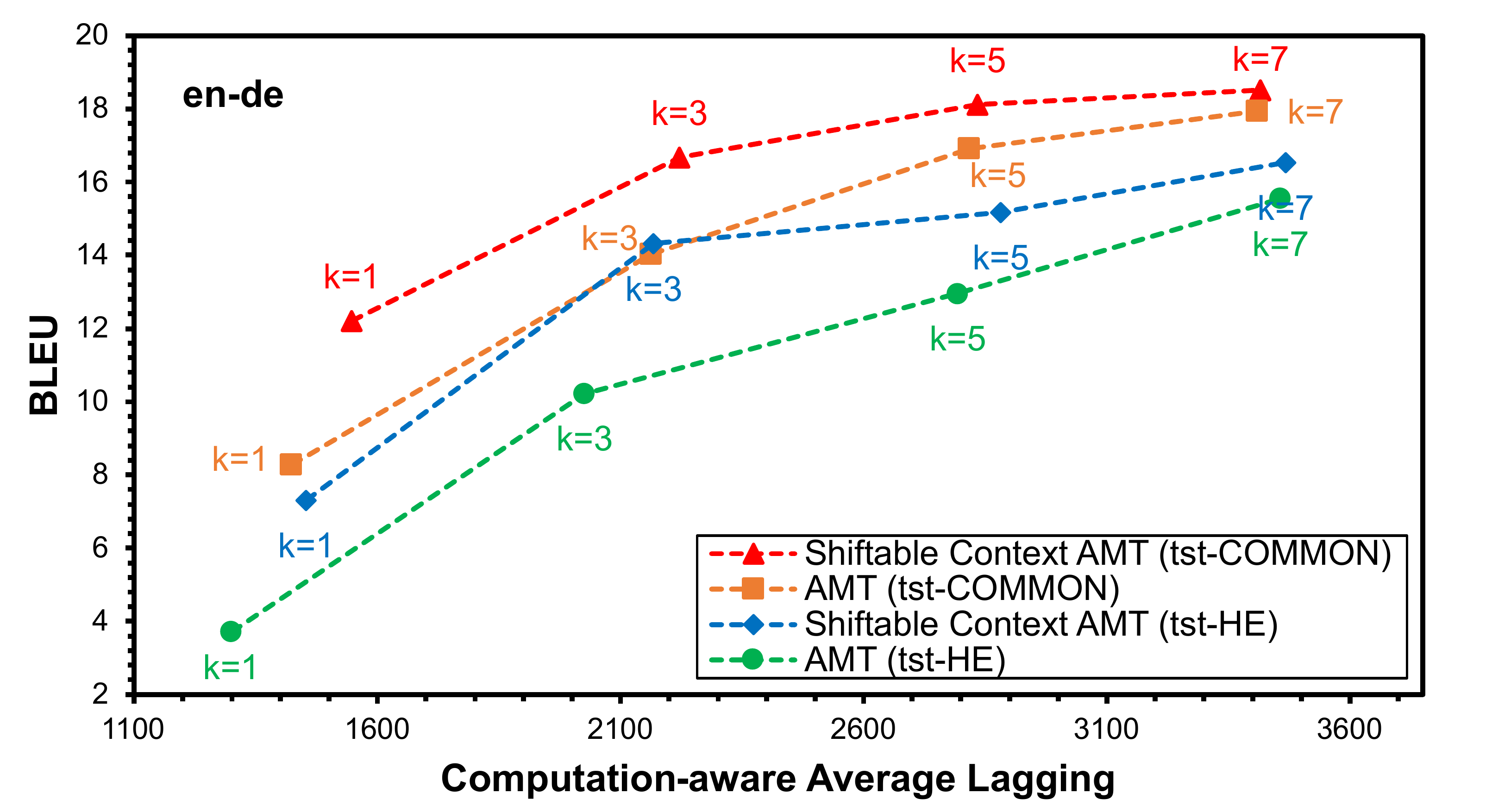}}
\vskip -0.1in
\caption{A plot comparing an Augmented Memory Transformer with and without the proposed shiftable context across four wait-$k$ values on the \textbf{en-de} language pair.}
\label{fig:en-decontext}
\end{center}
\vskip -0.2in
\end{figure}

\begin{figure}[ht]
\vskip 0.1in
\begin{center}
\centerline{\includegraphics[width=\columnwidth]{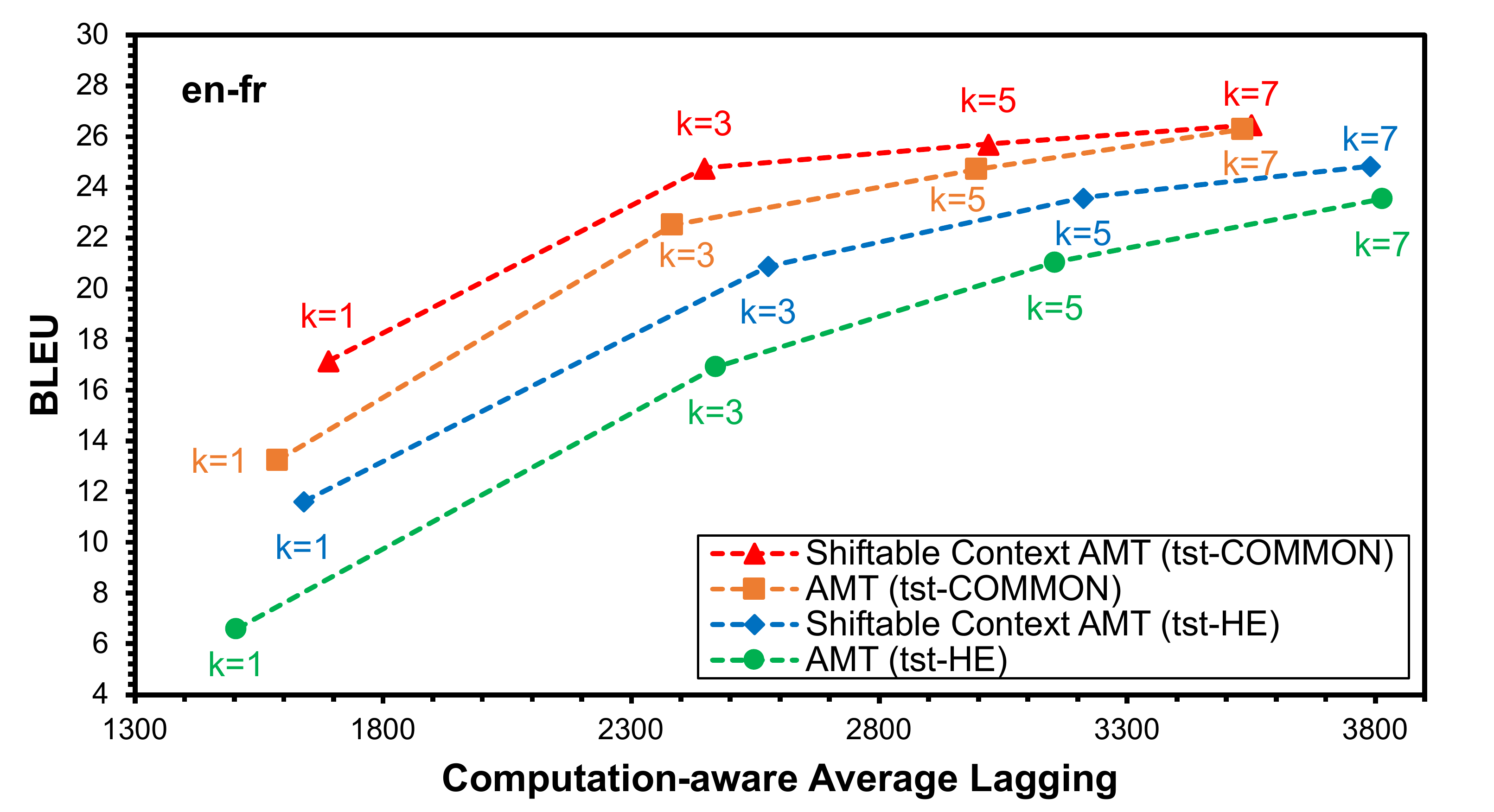}}
\caption{A plot comparing an Augmented Memory Transformer with and without the proposed shiftable context across four wait-$k$ values on the \textbf{en-fr} language pair.}
\label{fig:en-frcontext}
\end{center}
\vskip -0.2in
\end{figure}

\begin{figure}[ht]
\vskip 0.1in
\begin{center}
\centerline{\includegraphics[width=\columnwidth]{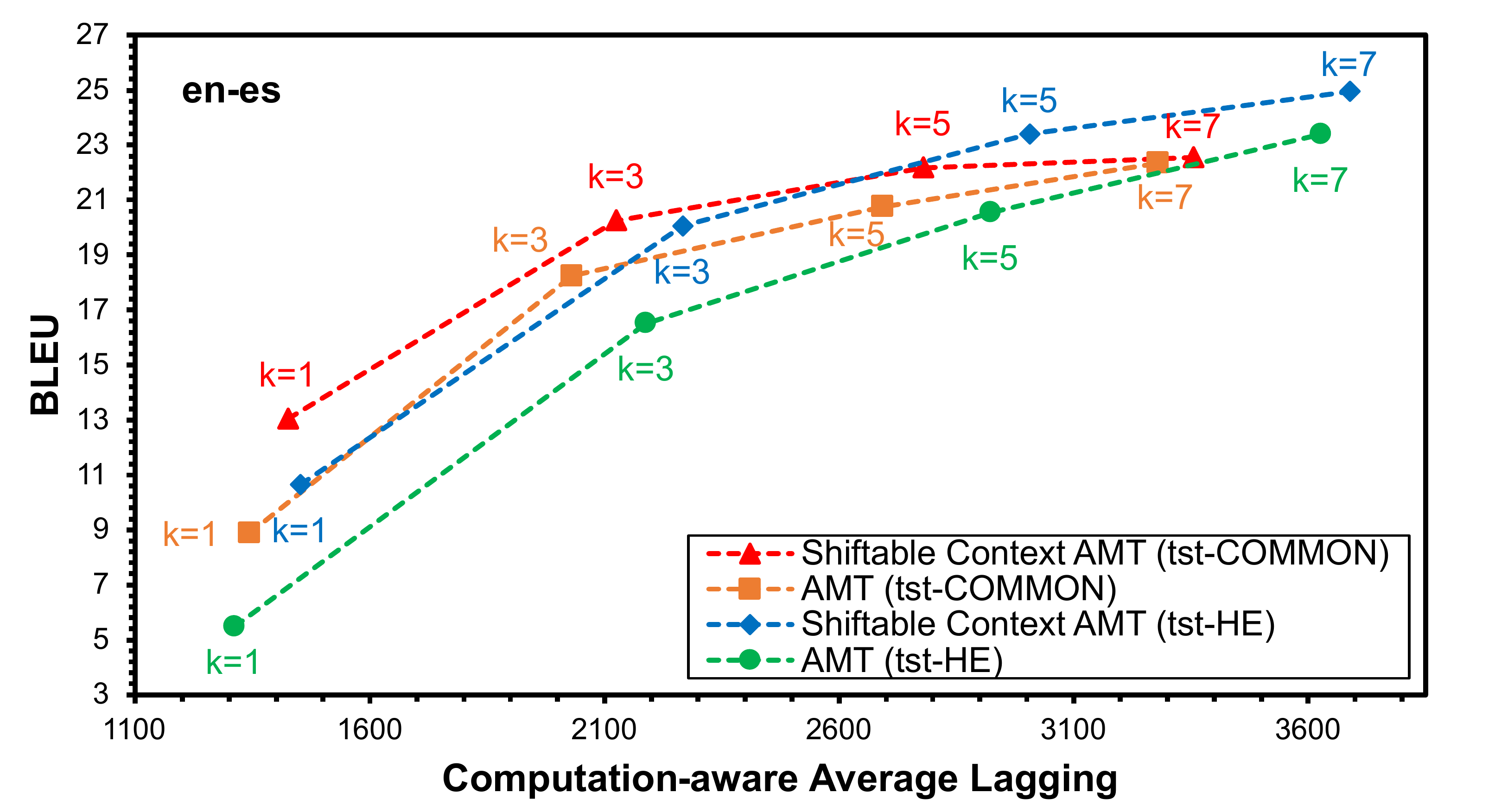}}
\caption{A plot comparing an Augmented Memory Transformer with and without the proposed shiftable context across four wait-$k$ values on the \textbf{en-es} language pair.}
\label{fig:en-escontext}
\end{center}
\vskip -0.2in
\end{figure}

We tested the combined effect of the shiftable left, center, and right context on an Augmented Memory Transformer by plotting its BLEU score and computation-aware Average Lagging across different wait-$k$ values.  We also plot the same configuration for an Augmented Memory Transformer, not using our techniques for comparison.  
The results for the English-German, English-French, and English-Spanish evaluations are shown in Figures \ref{fig:en-decontext}, \ref{fig:en-frcontext}, and \ref{fig:en-escontext} respectively.

From Figure \ref{fig:en-decontext}, we can see the shiftable left, center, and right context together provided an increase over the baseline Augmented Memory Transformer across all wait-$k$ values. The increase was, on average, 2.09 BLEU across each wait-$k$ value for the tst-COMMON test set and 2.72 BLEU for the tst-HE test set.  Furthermore, the computational increase required for completing partial sequence inputs had a minimal impact on the latency of the model, as seen by the computation-aware Average Lagging being inline for both models at each wait-$k$ value.  When measured, the increase in computation-aware Average Lagging was only 0.052 seconds for the tst-COMMON test set and 0.098 seconds for the tst-HE test set when averaged across all wait-$k$ values, which are minor compared with the 2-3 seconds base delay.  

Similar impressive results are visible in both Figure \ref{fig:en-frcontext} and \ref{fig:en-escontext}. 
In the case of the English-French models shown in Figure \ref{fig:en-frcontext}, our contributions provided an average 1.83 BLEU increase on the tst-COMMON test set and a 3.19 BLEU increase on the tst-HE test set across all wait-$k$ values.  Similarly, the BLEU score for the English-Spanish models shown in Figure \ref{fig:en-escontext} increased by 1.96 BLEU on the tst-COMMON test set and 3.28 BLEU on the tst-HE test set with our contributions.  In both cases, the additional computational overhead had a minimal impact on the computation-aware Average Lagging.

Another observation from the models trained on each dataset was that the accuracy improvement of the shiftable context is larger for smaller wait-$k$ values.  As explained in Section \ref{Sec:ShiftableLeftContext}, this is due to models with a lower wait-$k$ value experiencing the influence of the training-inference context mismatch for initial segments in a sentence to a greater degree.  Such a result is caused by models with larger wait-$k$ values having multiple complete segments by the time the simultaneous decoder begins translating, whereas lower wait-$k$ value models will need to overcome incomplete initial segments, which our shiftable context scheme overcomes.  As such, our contributions work especially well for low-latency scenarios achieving additional translation accuracy with minimal computational overhead. 
\subsection{Ablation Study}
\label{Sec:Ablation}
We performed an ablation study to identify the influence of the shiftable left, right, and center contexts individually.  All tests were performed on the English-German language pair of the MUST-C dataset.  
\begin{figure}[ht]
\vskip 0.1in
\begin{center}
\centerline{\includegraphics[width=\columnwidth]{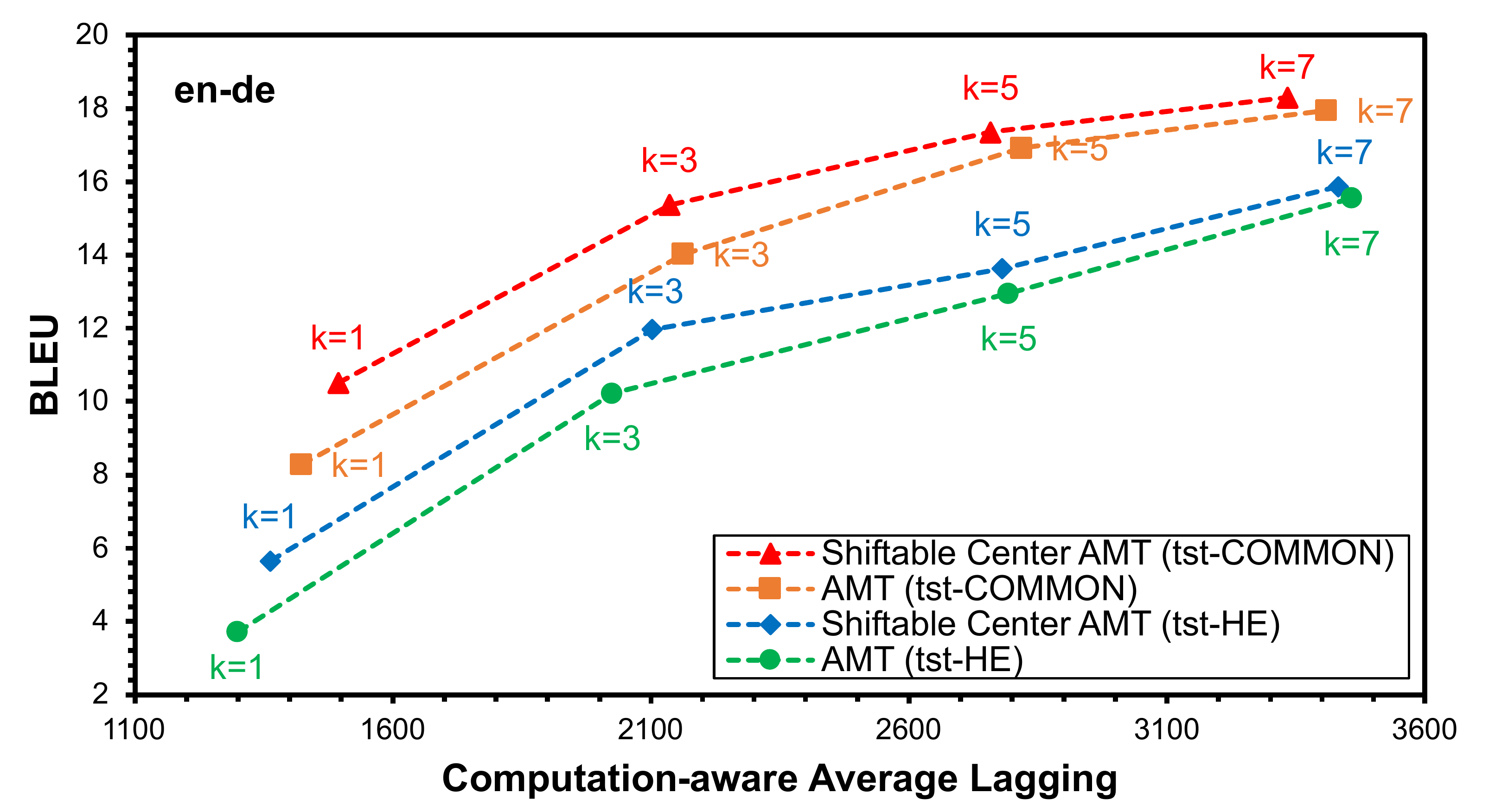}}
\caption{A comparison of an Augmented Memory Transformer with and without a shiftable \textbf{center} context on the \textbf{en-de} language pair using computation-aware Average Lagging and BLEU score.}
\label{fig:centeren-de}
\end{center}
\vskip -0.2in
\end{figure}

We show the results for the influence of the shiftable center context in Figure \ref{fig:centeren-de}. 
On average, across all wait-$k$ values, the performance increased by 1.10 BLEU for the tst-COMMON test set and 1.17 BLEU for the tst-HE test set.  Additionally, Figure \ref{fig:centeren-de} demonstrates the shiftable center context had a minimal effect on the computation-aware Average Lagging for each wait-$k$ value.

\begin{figure}[ht]
\vskip 0.1in
\begin{center}
\centerline{\includegraphics[width=\columnwidth]{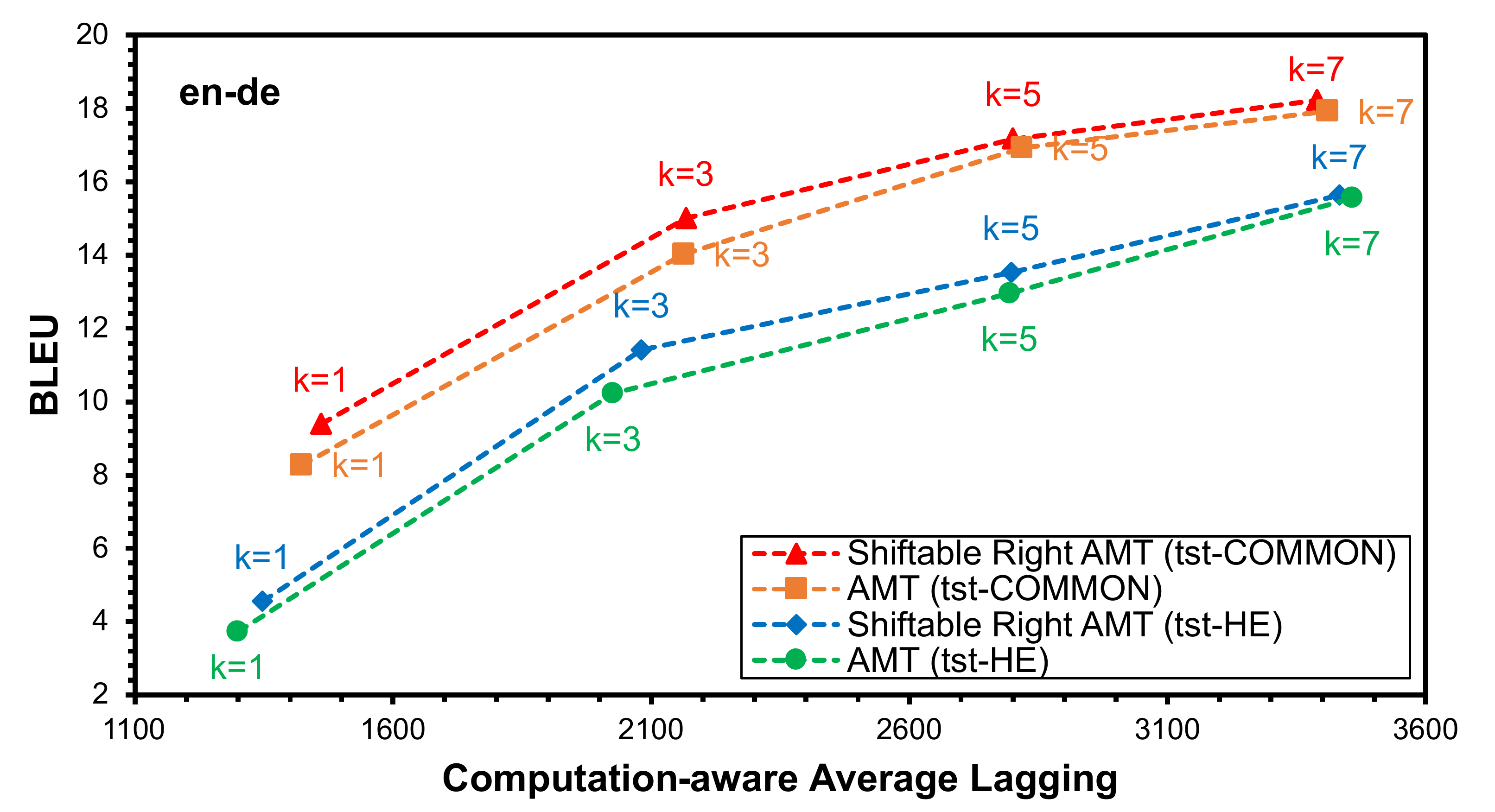}}
\caption{A comparison of an Augmented Memory Transformer with and without a shiftable \textbf{right} context on the \textbf{en-de} language pair using computation-aware Average Lagging and BLEU score.}
\label{fig:righten-de}
\end{center}
\vskip -0.2in
\end{figure}

The results for the shiftable right context are shown in Figure \ref{fig:righten-de}.  In this case, the increase in BLEU score from the shiftable right context was 0.66 BLEU on the tst-COMMON test set and 0.67 BLEU on the tst-HE test set when averaged across all wait-$k$ values.  As with the shiftable center context, the shiftable right context had a minimal influence on the computation-aware Average Lagging for each wait-$k$ value.

\begin{figure}[ht]
\vskip 0.1in
\begin{center}
\centerline{\includegraphics[width=\columnwidth]{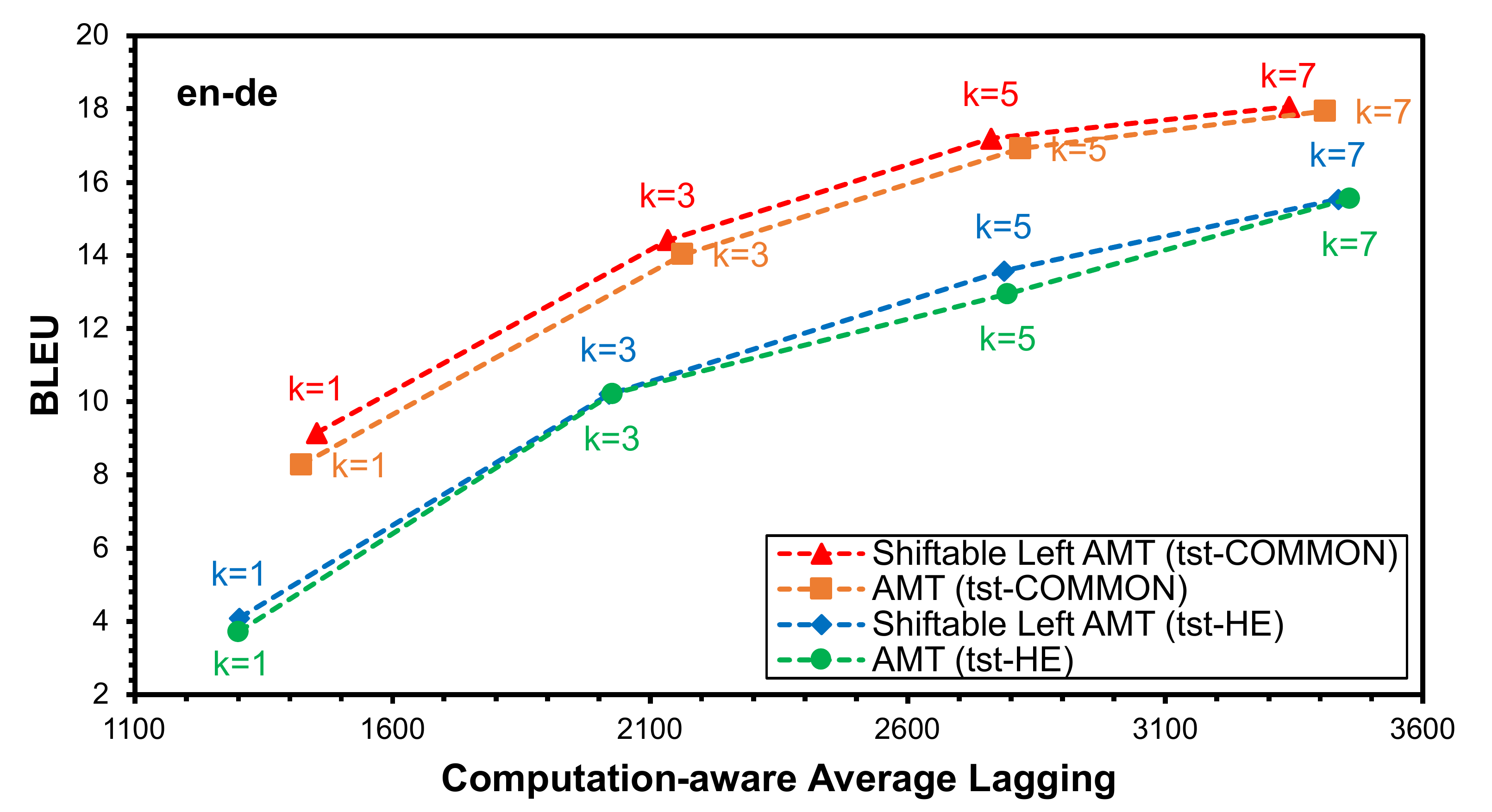}}
\caption{A comparison of an Augmented Memory Transformer with and without a shiftable \textbf{left} context on the \textbf{en-de} language pair using computation-aware Average Lagging and BLEU score.}
\label{fig:leften-de}
\end{center}
\vskip -0.2in
\end{figure}

Similar trends are also seen with the shiftable left context provided in Figure \ref{fig:leften-de}. The results in Figure \ref{fig:leften-de} show the Augmented Memory Transformer using the shiftable left context outperformed the baseline by 0.42 BLEU on the tst-COMMON test set and 0.23 BLEU on the tst-HE test set when averaged across all wait-$k$ values. As previously seen with the shiftable center and right context, the shiftable left context did not have a significant influence on the computation-aware Average Lagging. The results also confirm that improving the translation of the first segment alone leads to a sizable improvement in the overall BLEU score.

Provided the above improvement of the BLEU score for the shiftable left, center, and right context when isolated, each is a necessary component for solving the training-inference context mismatch.  Moreover, since each technique is effective, it presents the opportunity to improve the performance of alternative segment-based transformers whose segment composition may prevent them from using a combination of our techniques (i.e. a lack of right context).  

\section{Conclusion}
With the ever-increasing connectivity of the world, the difficult task of SimulST is becoming a necessity.  
Even with the solid performance of the state-of-the-art Augmented Memory Transformer, it still suffers from a mismatch between its training and inference environments, whereby the new chunk of tokens it receives by a given timestep, along with the previous partial input, is unable to completely fill the space allotted for each segment.  As such, we propose shiftable context, a scheme applicable to all segment-based transformers but demonstrated on the Augmented Memory Transformer to keep the size of segments constant during SimulST.  This scheme is composed of three techniques consisting of a shiftable left, center, and right context.  With these techniques combined, we are able to ensure that each segment reaches its normal size as long as there are available tokens not yet used from the input sequence.  By applying the proposed shiftable context to an Augmented Memory Transformer, we observed a 2.09, 1.83, and 1.95 BLEU increase for the English-German, English-French, and English-Spanish language pairs, respectively, on the tst-COMMON test set from the MUST-C dataset when averaged across wait-$k$ values, with a minimal impact on computation-aware Average Lagging.  In our ablation study isolating the contribution of each technique, sizable improvements were observed for every technique.  

\textbf{Limitations:} From the promising performance of our shiftable center, right, and left context on the Augmented Memory Transformer, there is a need to adapt such ideas to other segment-based transformers such as the Transformer-XL, Emformer, or Implicit Memory Transformer \cite{dai2019transformer, shi2021emformer, raffel-etal-2023-implicit}.  Furthermore, we were limited to applying our modified Augmented Memory Transformer strictly to SimulST when it is applicable to other streaming tasks, such as automatic speech recognition. The contributions from this paper move towards eliminating the need for human simultaneous interpreting, raising ethical concerns related to job security for this sector.  


\section*{Software and Data}
\label{Sec:Software}
Our publicly available implementation of the shiftable context for the Augmented Memory Transformer is provided in the following GitHub repository: \url{https://github.com/OSU-STARLAB/ShiftableContext}.

\section*{Acknowledgements}
This research was supported, in part, by the National Science Foundation grants 2223483 and 2223484.


\nocite{langley00}

\bibliography{example_paper}
\bibliographystyle{icml2023}

\newpage
\appendix
\onecolumn
\section{Examples}
\label{Appendix:Examples}
We provide examples showcasing the efficacy of our shiftable context scheme on the English-Spanish language pair of the tst-COMMON MuST-C test set.  Each example in Section \ref{Sec:Ex1}, \ref{Sec:Ex2}, \ref{Sec:Ex3}, and \ref{Sec:Ex4} contains the source speech transcript of the English audio and the respective target translation.  The audio file associated with the example is located in our GitHub repository.  Additionally, each example includes the predicted tokens, complete output, and BLEU score for both the baseline Augmented Memory Transformer and our Augmented Memory Transformer modified with shiftable context.  Each example uses a wait-5 policy with a pre-decision ratio of 8 (a chunk size of 32 tokens). The segment size in each example contains a left context of 32 tokens, a center context of 32 tokens, and a right context of 32 tokens.  The number of memory banks is 3.  We provide a guided explanation for Section \ref{Sec:Ex1}.
\subsection{Example 1}
\label{Sec:Ex1}
Suppose we want to translate the English speech for the below example to Spanish text. From the wait-5 policy, the model will wait for 5 chunks of 32 tokens or 160 tokens before beginning translation. Each token provides 10 milliseconds of new information. Therefore translation will begin after a delay of 1.6 seconds.  We will first begin by walking through the translation generation process using a baseline Augmented Memory Transformer followed by the same process with the shiftable context Augmented Memory Transformer to show the importance of shiftable context in practice.

In the baseline, we have a regular Augmented Memory Transformer with segments consisting of a left context of 32 tokens, a center context of 64 tokens, and a right context of 32 tokens (32+64+32). Such a segment contains 1.28 seconds of audio information. After receiving 1.6 seconds of audio or 160 tokens, the first, second, and third segments will have a representation of 0+64+32, 32+64+32, and 32+32+0. Once each segment is processed, the hidden states created from the center context are provided to the decoder. The decoder then produces the first output prediction vocab token “\_No”, which is correct for the target translation. When the next chunk of 32 tokens is received, the three segments created are 0+64+32, 32+64+32, and 0+64+32. The subsequent predicted vocab token is “\_sorprende”, which deviates from the accurate translation vocab token “\_es”. This deviation occurs because the hidden states associated with segments 0+64+32 and 32+64+0 have insufficient representations for the decoder to generate an accurate translation. Similarly, once the next chunk is received, the new segments created are 0+64+32, 32+64+32, 32+64+32, and 32+32+0. The respective token predicted is “esta”, which once again deviates from the target of “de”. In this case, the difference is due to the context mismatch; however, it also results from the autoregressive decoder misprediction in the previous time step. The compounding influence of the misprediction will continue for the subsequent predicted vocab tokens.

In contrast, the same Augmented Memory Transformer using shiftable context, after receiving the first 160 tokens, will create the segments 0+64+64, 32+64+32, and 96+32+0. The respective decoder token produced is “\_No”. When the next chunk is received, the segments created are 0+64+64, 32+64+32, and 64+64+0. This allows for the correct decoder prediction of “\_es”, which differs from the baseline Augmented Memory Transformer without shiftable context. Following this, when another new chunk is received, the subsequent segments created are 0+64+64, 32+64+32, 32+64+32, and 96+32+0. From the hidden states of these segments and the previous predictions, the decoder predicts the next vocab token as “\_de”, which is correct once again. Clearly, the shiftable context method reduces the deviation from the reference translation, contributing to an increased number of correct predictions.

\textbf{Source speech transcript:} ``Not surprisingly, this destruction also endangers bonobo survival."

\textbf{Target translation:} ``No es de (it is not) extrañar (surprising) que (that) esta (this) destrucción (destruction)  también (also) ponga (puts) en (in) peligro (danger) la (the) supervivencia (survival)  del (of the) bonobo."

\textbf{Predicted baseline tokens:} ``\_No", ``\_sorprende", ``\_esta",  ``\_destrucción", ``.", ``$<$/s$>$"

\textbf{Predicted baseline output:} ``No sorprende (No surprise) esta (this) destrucción (destruction). $<$/s$>$"

\textbf{BLEU score:} 5.21

\textbf{Predicted shiftable context tokens:} ``\_No", ``\_es", ``\_de", ``\_extraña", "r", ``\_que", ``\_esta", ``\_destrucción", ``\_también", ``\_sea", ``\_de", ``\_sobreviviente", ``s", ``\_de", ``\_Bo", ``n", ``na", ``va", ``.", ``$<$/s$>$"

\textbf{Predicted shiftable context output:} ``No es de (it is not) extrañar (surprising) que (that) esta (this) destrucción (destruction)  también (also) sea de (from) sobrevivientes (survivors) de (of) Bonnava. $<$/s$>$"

\textbf{BLEU score:} 40.05

\subsection{Example 2}
\label{Sec:Ex2}
\textbf{Source speech transcript:} ``I also believe that in many parts of this country, and certainly in many parts of this globe, that the opposite of poverty is not wealth. I don't believe that. I actually think, in too many places, the opposite of poverty is justice."

\textbf{Target translation:} ``También (Also) creo (I believe) que (that) en (in) muchas (many) partes (parts) de (of) este (this) país (country) y (and), sin duda (certainly), en (in) muchas (many) partes (parts) del (of the) mundo (globe), lo (the) opuesto (oposite) a (of) la (the) pobreza (poverty) no es (is not) la (the) riqueza (wealth). Así no es (It is not like that). En verdad (Actually) pienso (I think) que (that) en (in) muchas (many) partes (parts) lo (the) opuesto (opposite) a (of) la pobreza (poverty) es (is) la justicia (justice)."

\textbf{Predicted baseline tokens:}  ``\_También", ``\_creo", ``\_que", ``\_en", ``\_muchas", ``\_partes", ``\_de", ``\_este", ``\_país",  ``\_en", ``\_muchos", ``\_aspectos", ``\_de", ``\_este", ``\_mundo", ``.", ``$<$/s$>$"

\textbf{Predicted baseline output:} ``También (Also) creo (I believe) que en (that) muchas (many) partes (parts) de (of) este (this) país (country), en (in) muchos (many) aspectos (aspects) de (of) este (this) mundo (globe). $<$/s$>$"

\textbf{BLEU score:} 10.81

\textbf{Predicted shiftable context tokens:} ``\_También", ``\_creo", ``\_que", ``\_en", ``\_muchas", ``\_partes", ``\_de", ``\_este", ``\_país", ",", ``\_y", ``\_ciertamente", ``\_en", ``\_muchas", ``\_partes", ``\_de", ``\_este", ``\_mundo", ",", ``\_que", ``\_lo", ``\_contrario", ``\_de", ``\_la", ``\_pobreza", ``\_no", ``\_es", ``\_riqueza", ",", ``\_no", ``\_lo", ``\_creo", ".", ``\_De", ``\_hecho", "," ``\_creo", ``\_que", ``\_en", ``\_muchos", ``\_lugares", ",", ``\_lo", ``\_opuesto", ``\_a", ``\_la", ``\_pobreza", ``\_la", ``\_justicia", ``.", ``$<$/s$>$"

\textbf{Predicted shiftable context output:} ``También (Also) creo (I believe) que (that) en (in) muchas (many) partes (parts) de (of) este (this) país (country), y (and) ciertamente (certainly) en (in) muchas (many) partes (parts) de (of) este (this) mundo (globe), que lo contrario (the opposite) de (of) la pobreza (poverty) no es (is not) riqueza (wealth), no lo creo (I don’t believe that). De (In) hecho (fact), creo (I think) que (that) en (in) muchos (many) lugares (places), lo (the) opuesto (opposite) a (of) la pobreza (poverty) es (is) la justicia (justice). $<$/s$>$"

\textbf{BLEU score:} 40.38

\subsection{Example 3}
\label{Sec:Ex3}
\textbf{Source speech transcript:} ``A second possibility is that there will be evolution of the traditional kind, natural, imposed by the forces of nature."

\textbf{Target translation:} ``Una (A) segunda (second) posibilidad (possibility) es (is) que (that) se produzca (produces) una (an) evolución (evolution) del (of the) tipo tradicional (traditional kind), natural (natural), impuesta (imposed) por (by) las (the) fuerzas de la Naturaleza (forces of nature)."

\textbf{Predicted baseline tokens:} ``\_La", ``\_segunda", ``\_posibilidad", " \_es", ``\_que", ``\_habrá", ``\_una", ``\_evolución", ``\_del", ``\_tipo", ``\_tradicional", ``.", ``$<$/s$>$"
Predicted baseline output: ``La (The) segunda (second) posibilidad (possibility) es (is) que (that) habrá (there will be) una (an) evolución (evolution) del (of) tipo tradicional (traditional kind). $<$/s$>$"

\textbf{BLEU score:} 25.42

\textbf{Predicted shiftable context tokens:} ``\_La", ``\_segunda", ``\_posibilidad", ``\_es", ``\_que", ``\_habrá", ``\_una", ``\_evolución", ``\_del", ``\_tipo", ``\_tradicional", ``,", ``\_natural", ``,", ``\_imp", ``on", ``s", ``able", ``\_por", ``\_las" , ``\_fuerzas", ``\_de", ``\_la", ``\_naturaleza", ``.", ``$<$/s$>$"

\textbf{Predicted shiftable context output:} ``La (The) segunda (second) posibilidad (possibility) es (is) que (that) habrá (there will be) una (an) evolución (evolution) del (of the) tipo tradicional (traditional kind), natural (natural), imponsable (imposed) por (by) las (the) fuerzas de la naturaleza (forces of nature). $<$/s$>$"

\textbf{BLEU score:} 49.86

\subsection{Example 4}
\label{Sec:Ex4}
\textbf{Source speech transcript:} ``We see the same thing with the disability rights movement.

\textbf{Target translation:} ``Vemos (We see)  lo (the) mismo (same) con (with) el (the) Movimiento de la Discapacidad (Disability Movement)."

\textbf{Predicted baseline tokens:} ``\_Vemos", ``\_lo", ``\_mismo", ``\_con", ``\_la", ``\_mujer", ``\_de", ``\_derechos",  ``\_civiles", ``.", ``$<$/s$>$"

\textbf{Predicted baseline output:} ``Vemos (We see)  lo (the) mismo (same) con (with) la (the) mujer (woman) de (of) derechos civiles (civil rights). $<$/s$>$"

\textbf{BLEU score:} 20.45

\textbf{Predicted shiftable context tokens:} ``\_Vemos", ``\_lo", ``\_mismo", ``\_con", ``\_el", ``\_movimiento", ``\_de", ``\_derechos",  ``\_civiles", ``.", ``$<$/s$>$"

\textbf{Predicted shiftable context output:} ``Vemos (We see)  lo (the) mismo (same) con (with) el (the) movimiento (movement) de (of) derechos civiles (civil rights). $<$/s$>$"  

\textbf{BLEU score:} 28.92


\end{document}